# Implementation of Robust Face Recognition System Using Live Video Feed Based on CNN

Yang Li and Sangwhan Cha

Computer and Information Science
Harrisburg University of Science and Technology, PA, USA
{yli, scha}@harrisburgu.edu

## ABSTRACT

The way to accurately and effectively identify people has always been an interesting topic in research and industry. With the rapid development of artificial intelligence in recent years, facial recognition gains lots of attention due to prompting the development of emerging identification methods. Compared to traditional card recognition, fingerprint recognition and iris recognition, face recognition has many advantages including non-contact interface, high concurrency, and user-friendly usage. It has high potential to be used in government, public facilities, security, e-commerce, retailing, education and many other fields. With the development of deep learning and the introduction of deep convolutional neural networks, the accuracy and speed of face recognition have made great strides. However, the results from different networks and models are very different with different system architecture. Furthermore, it could take significant amount of data storage space and data processing time for the face recognition system with video feed, if the system stores images and features of human faces. In this paper, facial features are extracted by merging and comparing multiple models, and then a deep neural network is constructed to train and construct the combined features. In this way, the advantages of multiple models can be combined to mention the recognition accuracy. After getting a model with high accuracy, we build a product model. The model will take a human face image and extract it into a vector. Then the distance between vectors are compared to determine if two faces on different picture belongs to the same person. The proposed approach reduces data storage space and data processing time for the face recognition system with video feed scientifically with our proposed system architecture.

## 1. INTRODUCTION

Computers and information technologies are rapidly integrating into everyday human life. Face recognition is one of such technologies and gains lots of attentions due to prompting the development of emerging identification methods. As the digital world and real-world merge more and more together, how to accurately and effectively identify users and improve information security has become an important research topic. Different from the traditional identity recognition technology, biometrics is the use of the inherent characteristics of the body for identification such as fingerprints, irises and face[1].

Some traditional face recognition algorithms identify facial features by extracting landmarks, or features, from an image of the subject's face. For example, algorithm may analyze the relative position, size, and/or shape of the eyes, nose, cheekbones, and jaw. These features are then used to search for other images with matching features [2]. These kinds of algorithms can be complicated, require lots of compute power, hence could be slow in performance. And they can also be inaccurate when the faces show clear emotional expressions, since the size and position of the landmarks can be altered significantly in such circumstance.

High accuracy face recognition models have been reported in scientific researches by giant technology companies and research institutions, as shown in Table 1 [3]. However, all of these ground-breaking result remain in the laboratory. The development of face recognition applications in the real world could be significantly complicated since there are many factors to consider. For example, FaceNet is able to work well for identifying fake ID while it is not well fitted for identifying the right person who picks up the right kid in a daycare.

Table 1. Face recognition models and their accuracy

| Method | Net. Loss | Outside data | # models | Aligned | Verif. metric | Layers | Accu. |
|---|---|---|---|---|---|---|---|
| DeepFace [97] | ident. | 4M | 4 | 3D | wt. chi-sq. | 8 | 97.35±0.25 |
| Canon. view CNN [115] | ident. | 203K | 60 | 2D | Jt. Bayes | 7 | 96.45±0.25 |
| DeepID [92] | ident. | 203K | 60 | 2D | Jt. Bayes | 7 | 97.45±0.26 |
| DeepID2 [88] | ident. + verif. | 203K | 25 | 2D | Jt. Bayes | 7 | 99.15±0.13 |
| DeepID2+ [93] | ident. + verif. | 290K | 25 | 2D | Jt. Bayes | 7 | 99.47±0.12 |
| DeepID3 [89] | ident. + verif. | 290K | 25 | 2D | Jt. Bayes | 10-15 | 99.53±0.10 |
| Face++ [113] | ident. | 5M | 1 | 2D | L2 | 10 | 99.50±0.36 |
| FaceNet [82] | verif. (triplet) | 260M | 1 | no | L2 | 22 | 99.60±0.09 |
| Tencent [8] | - | 1M | 20 | yes | Jt. Bayes | 12 | 99.65±0.25 |

The main gap between face recognition research and industrial usage comes from the way to develop applications based on algorithm APIs. Since there are many different APIs provided by different companies, it would be difficult how to turn the APIs into the real product for potential industrial users.

In this paper, we propose to build a high performance, scalable, agile, and low-cost face recognition system. Based on the theory of deep learning, we build the Siamese network which will train the neural network based on similarities. Once we examine and compare the available open source data set, we chose AT&T "The database of faces", which is formerly known as ORL dataset and trained the model with GPU. The model will take a human face image and extract it into a vector. Then the distance between vectors are compared to determine if two faces on different picture belongs to the same person. This approach reduces data storage space and data processing time for the face recognition system with video feed scientifically with our proposed system architecture.

The remaining of this paper is organized as follows. Section 2 presents related work. Section 3 describes the proposed approach. Building a facial recognition model is described in Section 4. Section 5 provides the conclusion.

## 2. RELATED WORK

The author in [4] proposed a robust real-time object detection, which made a face detection truly feasible for various face recognition systems that have been devised and applied over the past several years of technological development [1,5,6]. The main goal of face recognition system is to find the position and size of each face in the image or video, but for tracking, it is also necessary to determine the correspondence between different faces in the frame.

Earlier localization of facial feature points focused on two or three key points, such as locating the center of the eyeball and the center of the mouth, but later introduced more points and added mutual restraint to improve the accuracy and stability of positioning. The paper that is titled "Active shape models-their training and application" [7] has been a model of dozens of facial feature points and texture and positional relationship constraints considered together for calculation. The regression-based approach presented in [8] shows better results than the approaches based on the categorical apparent model. The paper that is titled "Face alignment by explicit shape regression" [9] presented another aspect of Active Shape Model (ASM) improvement and an improvement on the shape model that is based on the linear combination of training samples to constrain the shape and the effect of alignment.

The purpose of the facial feature point positioning is to further determine facial feature points (eyes, mouth center points, eyes, mouth contour points, organ contour points, etc.) on the basis of the face area detected by the face detection, tracking and position. These researches [7,8,9] show the methods for face positioning and face alignment. The basic idea of locating the face feature points is to combine the texture features of the face locals and the position constraints of the organ feature points.

There are researches [10,11,12] that discussed about facial feature positioning and alignment. Facial feature extraction is a face image into the string of fixed-length numerical process. This string of numbers is called the "Face Feature" and has the ability to characterize this face. Human face to mention the characteristics of the process of input is "a face map" and "facial features key points coordinates". The output is the corresponding face of a numerical string (feature). Face to face feature algorithm will be based on facial features of the key point coordinates of the human face pre-determined mode, and then calculate the features. In recent years, the deep learning algorithms basically ruled the face lift feature algorithm. In [10,11,12], they showed the progress of research in this area by presenting fixed time length algorithms. Earlier face feature models are large and slow which could be only used in the background service. However, some recent studies can optimize the model size and operation speed to be available to the mobile terminal under the premise of the basic guaranteed algorithm effect.

## 3. BUILDING PROPOSED SYSTEM

Many aspects need to be taken into consideration when building towards a commercial usable system. We are targeting at a system architecture with high performance, scalability, agility, and low costs. The high performance means the system will give out result at milliseconds level and have high threshold with high concurrency. Scalability means the system can be scaled out well as the need increases from a single node machine to a multi node cluster. Agility means the system should be easily modifiable and be able to apply to different domain. Low costs mean low development costs as well as development

and maintenance costs. To achieve all the goals, we need to resolve many practical problems

## CHOOSE BETWEEN CPU AND GPU

Although GPU was not designed for neural network initially, but it has the features which put it into a better position than CPU for the neural network calculations: a) Provides the infrastructure of multi-core parallel computing and has a large number of cores which can support parallel computing of large amounts of data. Parallel computing is relative to serial computing. It is an algorithm that can execute multiple instructions at a time, with the goal of increasing the speed of calculations and solving large and complex computational problems by expanding the problem solving scale. b) Has a higher speed of memory access. c) Has higher floating point computing power. Floating-point computing power is an important indicator of multimedia and 3D graphics processing related to processors. In today's computer technology, due to the application of a large number of multimedia technologies, the calculation of floating point numbers has been greatly increased, such as the rendering of 3D graphics, so the ability of floating point computing is an important indicator to examine the computing power of the processor.

A test done by V Chu [13] shows that, depending on the GPU you choose, the performance gain of a GPU over CPU on neural network calculation can be between 43 times to 167 times faster. Therefore, to achieve the high performance, our system will need to use GPU for the neural network related processes.

## CHOOSE BETWEEN ALL-IN-ONE AND CLIENT-SERVER ARCHITECTURE

The demos from many face recognition research institutions are using all-in-one architecture. In these examples [18], all processing unit are inside same system. This is very good for prototyping but has issues on industrial applications. The first problem to be considered is the size, cost and power consumption of a graphic card. For example, if we are using a RTX 2080 Ti graphic card, the size is too big to be put into small size appliances. The power consumption is 280W, with huge amount of heats generated when running at full potential. The price is no less than $999 for one piece, which limit bulk deployment of it. Despite all the limitations of the graphic card, its computational power is superior. In our stress test with a GTX 1080 Ti graphic card, it can extract facial features from 200 pictures at same time with an average of 25 milliseconds processing time per picture. All the limitations and features are implicating that the server-client architecture will perform well.

## THE CLIENT-SERVER ARCHITECTURE WITH GPU

Figure 1 shows the client-server architecture with GPU we designed to fulfill the requirement. The whole system has 2 sub-systems: user registration system and real time recognition system, and these 2 sub-systems share compute-heavy components to reduce the complexity and cost. The client side of user registration system will take user picture and user input, i.e. employee id. Then the picture and user input are sent to the server, where GPU resides. The GPU will process the picture by detecting human face and extracting the picture into vector that is stored into database with the user input.

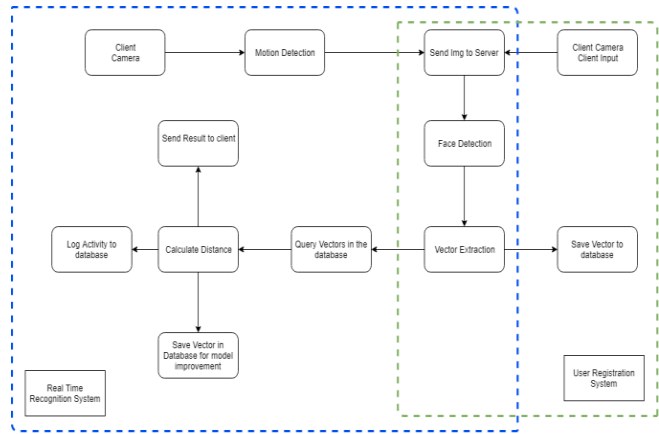

Figure 1. Client-server architecture with GPU

In the real time recognition system, the client camera will be used to capture real time videos. In our first design, the client will send video feed to the server whatever was captured. During the stress test, we found this method will load significant amount of video feed into the server. Therefore, we added a motion detection component to the client. This component can be developed with opencv libraries and can be deployed on both desk-top level clients and mobile clients. The motion detection will significantly reduce the amount of image which need to be processed for face detection.

After a motion is detected, the video feed will be processed and only send n frames of pictures to the server. The face detection components in the server will try to detect the face from the image, it will be able to detect multiple faces from single image. If there is no face detected, the image will be discarded. If one or multiple faces were detected, each face will be cropped from the original image, aligned, resized, and marked with a unique ID and send to the vector extraction processer. The vector extraction processer will read the pre-processed human face image and extract each face into a vector. The vector will be combined with the unique request ID and start a query into the face feature database. The Euclidean distance will be calculated against the current face vector and the face vectors retrieved from the database. This step can be done by GPU. After all the distances have been calculated, they will be sorted and generated the list n distances. The sample message generated is as follows.

```
{
        request_id:<x>,
    {
            user_id[1]:<x>
            distance:<x>
    },
    {
            user_id[2]:<x>
            distance:<x>
    },
    {
            user_id[3]:<x>
            distance:<x>
    }
}
```

This message will be sent back to the client to process its own rules on what will happen such as allowing employee badge-in, telling if an I.D. card is fake and etc. In the meanwhile, the message can be sent to a different database, which logs the activities, and it can be connected and configured with customers' ERP system. The images can also be saved for further training to improve the model performance.

**DATABASE DESIGN**

To preserve data, our proposed system will use its own database. We will use MySql relational database management system as an example in this paper. An obvious way for storing the human face picture into the database is to use blob (binary large object). However, there will be inefficiency in terms of space and time complexity. The latency would be very long for large dataset since the GPU needs to re-extract the vectors from the images of human faces stored in the databased, when the system performs the face comparison. To avoid this issue, we will first extract the vectors from human face image, and only save the vectors into the database. And we also need to associate the vector with an identification property. We decided to have User_id/employee_id/customer_id as an identification property. As shown in Table 2, the main table structure is depending on how many vectors the model extracts.

Table 2. Table Structure

| ID | Vector1 | Vector2 | ... | Vector<n> |
|---|---|---|---|---|

According to database normalization rules, the database belongs to this system should not store anything other than the ID and face vector. Since many user information could be changed rapidly in an organization, bringing any other column into this database will be redundant and require additional work to maintain data integrity. The database can retrieve/send information to other enterprise database like ERP, Clarify and etc. with just ID column as shown in Figure 2.

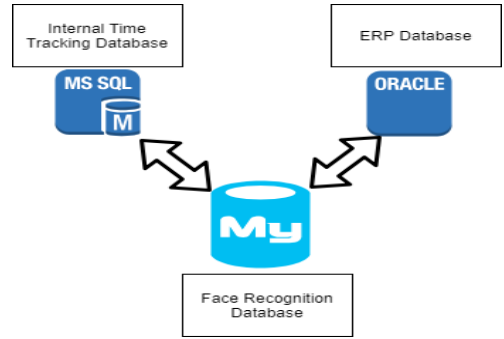

Figure 2. Database linking to each other

According to our experiment, one tweak in the database design is not to make the ID column a unique key or primary key column. By doing this, the performance can be significantly improved. For example, after the system is online for 10 days, we can gather one face with ID=9 10 times with slightly different vectors. We can save all these 10 vectors into the main table. When person with ID=10 go through the camera on the 11$^{th}$ day, the picture is processed and the top 3 candidates from the processing engine will show ID=10 more than 1 time. This result will give client more confidence to justify that the person is the one with ID=10.

**NODE FRAMEWORK AS SERVER**

Node.js is a JavaScript runtime environment, released in May 2009 by Ryan Dahl, which essentially encapsulates the Chrome V8 engine. Node.js is neither a JavaScript framework, nor a browser-side library. Node.js is a development platform that lets JavaScript run on the server side, making JavaScript a scripting language that is on par with server-side languages like PHP, Python, Perl, and Ruby.

The V8 engine itself uses some of the latest compilation techniques. This allows code written in a scripting language such as JavaScript to run at a much faster speed and saves development costs. The demand for performance is a key factor in Node. JavaScript is an event-driven language, and Node takes advantage of this to write highly scalable servers. Node uses an architecture called an "event loop" that makes writing a highly scalable server easy and secure. There are many different techniques for improving server performance. Node chose an architecture that improves performance while reducing development complexity. This is a very important feature. Concurrent programming is often complex and full of mines. Node bypasses these, but still provides good performance.

Node uses a series of "non-blocking" libraries to support the way the event loops. Essentially, it provides an interface for resources such as file systems and databases. When a request is sent to the file system, there is no need to wait for the hard disk (addressing and retrieving the file), and the non-blocking interface notifies Node when the hard disk is ready. The model simplifies access to slow resources in an extensible way, intuitive and easy to understand. Especially for users who are familiar with DOM events such as onmouseover and onclick, there is a feeling of deja vu.

Although running JavaScript on the server side is not unique to Node, it is a powerful feature. We found that the browser environment limits our freedom to choose a programming language. The desire to share code between any server and an increasingly complex browser client application can only be achieved through JavaScript. Although there are other platforms that support JavaScript running on the server side, because of the above characteristics, Node has developed rapidly and become top choice for many developers.

Figure 3 illustrates the diagram of how callback works in Node. Although the I/O does not cause the bottle neck in the face recognition system, Node still is the best candidate because for GPU to detect face and extract vector from face image and average of 25 milliseconds will be consumed. With the non-block feature of Node, the whole process will not just wait for 25 seconds each time a request comes. The Node sever will send the request to GPU, then keep handling new requests. After GPU finishes its work, it will run the call back function, and the server will pick up the next on the request and move one. It increases the concurrency of the system significantly.

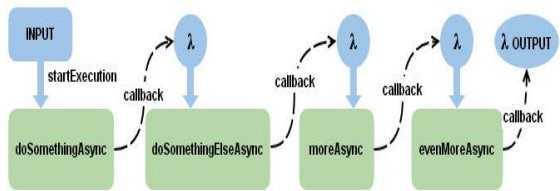

Figure 3. Callback in Node.js [14]

Figure 4 shows how the node server reacts with the neural network model in our proposed system. When the request comes in, the node sever will send the request to the neural network model. The neural network, which can be deployed in any format, will take the task and run it on the GPU. While this workload is running, node server will not wait for it to finish, but keep accepting requests from the caller. After the neural network model returns the calculation result, node server will package it with other necessary information and send the feedback to the caller.

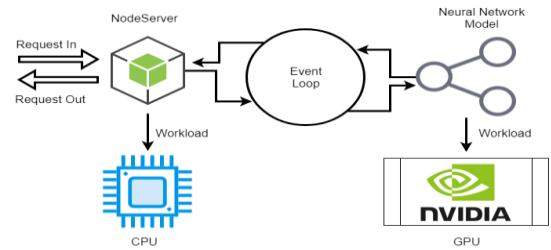

Figure 4. Node Server work with Neural Network Model

## 4. BUILDING FACE RECOGNITION MODEL

### BUILD SIAMESE NETWORK WITH CNN

After comparing different neural networks and their characteristics, we decided to use Siamese network to resolve the problem because of more robustness to class extreme imbalance. The Siamese network is neural network for measuring of similarity. It can be used for category identification and classification in the scenario when there are many categories, but the number of samples per category is small. The traditional classification method for distinguishing is to know exactly which class each sample belongs to and need to have an exact label for each sample. And the relative number of tags is not significant. These methods are less applicable when the number of categories is too large and the number of samples per category is relatively small. For the entire data set, our data volume is available to train. However, there could be only few samples for each category, which cause an unreliable result.

The Siamese network learns a similarity measure from the data and uses the learned metric to compare and match the samples of the new unknown category. This method can be applied to classification problems where the number of classes is large, or the entire training sample cannot be used for previous method training.

Our experiments have been performed on Ubuntu 18 operating system. The CPU is Intel(R) Core(TM) i5-7300HQ CPU @ 2.50GHz with 4 cores. The memory is dual channel DDR 4 8GB SDRAM. We train our neural network using NVIDIA GPU GeForce GTX 1050 with 4GB GPU RAM. The GPU driver version is 390.48. We use Compute Unified Device Architecture (CUDA) version 9.0, and NVIDIA CUDA Deep Neural Network (cuDNN) version 7.0 for CUDA 9.0. We build the Siamese network with below parameters:

```
SiameseNetwork(
  (cnn1): Sequential(
    (0): ReflectionPad2d((1, 1, 1, 1))
    (1): Conv2d(1, 4, kernel_size=(3, 3), stride=(1, 1))
    (2): ReLU(inplace)
    (3): BatchNorm2d(4, eps=1e-05, momentum=0.1, affine=True, track_running_stats=True)
    (4): ReflectionPad2d((1, 1, 1, 1))
    (5): Conv2d(4, 8, kernel_size=(3, 3), stride=(1, 1))
    (6): ReLU(inplace)
    (7): BatchNorm2d(8, eps=1e-05, momentum=0.1, affine=True, track_running_stats=True)
    (8): ReflectionPad2d((1, 1, 1, 1))
    (9): Conv2d(8, 8, kernel_size=(3, 3), stride=(1, 1))
    (10): ReLU(inplace)
    (11): BatchNorm2d(8, eps=1e-05, momentum=0.1, affine=True, track_running_stats=True)
  )
  (fc1): Sequential(
    (0): Linear(in_features=80000, out_features=500, bias=True)
    (1): ReLU(inplace)
    (2): Linear(in_features=500, out_features=500, bias=True)
    (3): ReLU(inplace)
    (4): Linear(in_features=500, out_features=5, bias=True)
  )
)
```

## TRAIN THE NEURAL NETWORK

The database for face recognition we choose is ORL. The ORL face database consists of 400 pictures of 40 people that is 10 pictures per person. The face has expressions, tiny gestures and so on. The training processing is performed on the two databases, and 90% of the faces in the databases are randomly selected as the training set, and the remaining 10% of the faces are used as test sets, and then the faces in the two sets are standardized. The training process was using GPU, as shown in figure 5 and 6. We found that the GPU usage went to 100% and the working temperature increased dramatically during training.

Figure 5. GPU Usage before training

Figure 6. GPU usage during training

We trained the model with 100 epochs. We found that the training loss goes down significantly during the early epochs, and converged to 0.0067 at last. Figure 7 shows clearly how the trend of training loss goes down as the epoch increases.

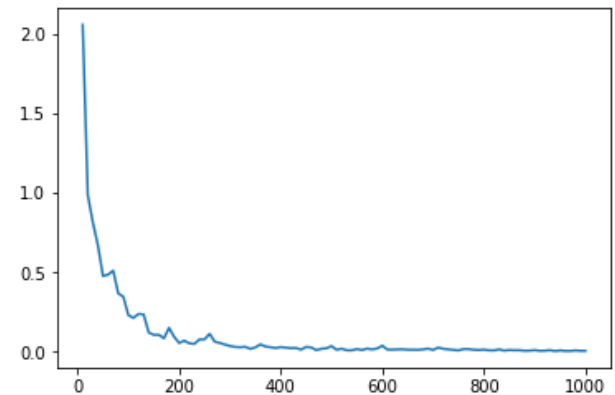

Figure 7. Training loss vs. Epoch

## MODEL VERIFICATION

The input of the neural network is an image of human face, and the output of the neural network is the vector of 5 dimensions. A sample output looks like below:

```
Vector of Face 1: Variable containing:
 1.7350  0.2165  1.0214  1.5764  2.2253
```

```
[torch.cuda.FloatTensor of size 1x5 (GPU 0)]
```

To calculate if the faces on two images come from the same person, we need to calculate the similarity of the two images, aka, the Euclidean distance between two vectors. Below is an example output of different people identified by our model, and the images are shown in Figure 8:

```
Vector of Face 1: Variable containing:
 1.7350  0.2165  1.0214  1.5764  2.2253
[torch.cuda.FloatTensor of size 1x5 (GPU 0)]
Vector of Face 2: Variable containing:
-0.7570  1.5081  0.3380  1.5524 -0.0977
[torch.cuda.FloatTensor of size 1x5 (GPU 0)]
Distance between Face1 Vector and Face2
Vector: Variable containing:
 3.7070
[torch.cuda.FloatTensor of size 1x1 (GPU 0)]
```

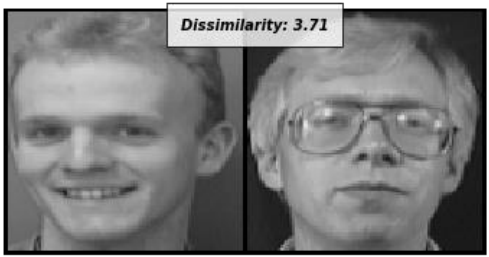

Figure 8. Different people with high Euclidean distance

The following is an example output of same person with different pose, identified by our model, and the images are shown in Figure 9:

```
Vector of Face 1: Variable containing:
 1.7350  0.2165  1.0214  1.5764  2.2253
[torch.cuda.FloatTensor of size 1x5 (GPU 0)]
Vector of Face 2: Variable containing:
 1.6301  0.7585  1.1658  1.6345  2.2486
[torch.cuda.FloatTensor of size 1x5 (GPU 0)]
Distance between Face1 Vector and Face2
Vector: Variable containing:
 0.5741
[torch.cuda.FloatTensor of size 1x1 (GPU 0)]
```

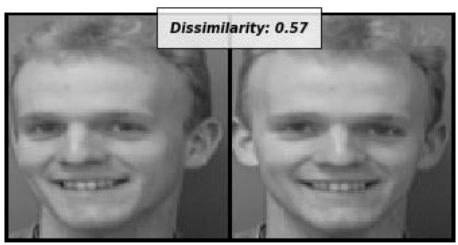

Figure 9. Same person with different pose, with low Euclidean distance

Based on our experiments, the Euclidean distance was higher than 3.00 for different people while it was lower than 1.00 for same people with different pose. However, it could be different values based on the size of data set we train the model.

## 4. CONCLUDING REMARKS

We propose to build a high performance, scalable, agile, and low-cost face recognition system. Based on the theory of deep learning, we build the Siamese network which will train the neural network based on similarities. Once we examine and compare the available open source data set, we chose ORL dataset and trained the model with GPU. The model will take a human face image and extract it into a vector. Then the distance between vectors are compared to determine if two faces on different picture belongs to the same person.

Then we compare, design and build a system to work with the neural network model. The system uses client-server architecture. GPU is used on the server side to provide high performance. We also de-coupled the main components of the system to make it flexible and scalable. We use the non-block and asynchronies features of Node.JS to increase the system's concurrency. Since the entire system is modularized, it can be used in different domains, which will cause reduced the development cost.

When we build the neural network model, there are many parameters which can be tuned to increase the model performance. We can keep tuning our models to increase its accuracy. Moreover, for a trained base model, we can re-train it using a specific dataset. Therefore, another way to increase the whole performance of system is to capture the specific people's images and re-train the model based on this small dataset. For example, if an organization with 3000 people uses this system, the model can be trained to be very accurate on these 3000 people. We can employ and automate this feature into the system.